\documentclass{article}

\usepackage[utf8]{inputenc}
\usepackage[T1]{fontenc}
\usepackage{amsmath}
\usepackage{amsfonts}
\usepackage{amssymb}
\usepackage{graphicx}
\usepackage{booktabs}
\usepackage{multirow}
\usepackage{longtable}
\usepackage{xcolor}
\usepackage{hyperref}
\usepackage{natbib}
\usepackage{caption}
\usepackage{subcaption}
\usepackage{algorithm}
\usepackage{algorithmic}

\usepackage[margin=1in]{geometry}
\usepackage{times}

\title{SCALPEL: Selective Capability Ablation via Low-rank Parameter Editing for Large Language Model Interpretability Analysis}

\author{
    Zihao Fu\\
    The Chinese University of Hong Kong\\
    \texttt{zihaofu@cuhk.edu.hk}
    \and
    Xufeng Duan\\
    The Chinese University of Hong Kong\\
    \texttt{xufengduan@cuhk.edu.hk}
    \and
    Zhenguang G. Cai\\
    The Chinese University of Hong Kong\\
    \texttt{zhenguangcai@cuhk.edu.hk}
}

\date{}

\begin{document}

\maketitle

\begin{abstract}
Large language models have achieved remarkable success across diverse domains, yet their deployment in many applications such as healthcare, legal systems, and autonomous decision-making remains limited by our incomplete understanding of their internal mechanisms. As these models become increasingly integrated into high-stakes systems, understanding how they encode and execute specific capabilities has become fundamental to interpretability research. Traditional approaches identify important modules through gradient attribution or activation analysis, assuming that specific capabilities are controlled by specific components. However, this assumption oversimplifies neural computation: individual modules may contribute to multiple capabilities simultaneously, and conversely, a single capability may be implemented in a distributed manner across multiple modules. These coarse-grained, module-level analyses fail to capture the fine-grained, distributed nature of capability encoding in neural networks. We present SCALPEL (Selective Capability Ablation via Low-rank Parameter Editing for Large language models), a framework that represents capabilities as low-rank parameter subspaces rather than discrete modules. Our key insight is that language model capabilities can be characterized by low-rank modifications distributed across layers and modules, enabling precise capability removal without affecting others. By training LoRA adapters to reduce the model's ability to distinguish correct from incorrect answers while preserving general language modeling quality, SCALPEL identifies the low-rank representation responsible for a particular capability while remaining disentangled from other capabilities. Experiments across diverse capability tasks and linguistic tasks from BLiMP demonstrate that SCALPEL successfully removes target capabilities while preserving other general capabilities, and provides fine-grained insights into how capabilities are distributed across the model's parameter space. Our results reveal that capabilities exhibit low-rank structure and can be selectively ablated through targeted parameter-space interventions, offering a more nuanced understanding of capability encoding in large language models.
\end{abstract}

\section{Introduction}

Large language models (LLMs)~\cite{openai2023gpt4,geminiteam2023gemini,zhao2023llmsurvey} have achieved remarkable success across diverse applications, from code generation~\cite{chen2021codex} to medical diagnosis~\cite{singhal2022medpalm} and scientific reasoning~\cite{zheng2023scientific}. However, their deployment in many applications such as healthcare~\cite{singhal2022medpalm}, legal systems~\cite{shui2023legal}, and autonomous decision-making~\cite{sha2023languagempc} remains limited by our incomplete understanding of their internal mechanisms. Without understanding how these models encode and process information, we cannot fully trust their decisions in applications requiring accountability. This opacity limits deployment in domains where interpretability and reliability are paramount.

To address these concerns, the interpretability research community has developed numerous approaches to understand how LLMs encode and process information. Gradient-based attribution methods~\cite{sundararajan2017axiomatic,selvaraju2017gradcam} identify which input features influence predictions, while activation analysis techniques~\cite{rimsky2024steering,nostalgebraist2020logitlens} reveal important components by examining hidden representations. Mechanistic interpretability methods~\cite{elhage2021mathematical,meng2022locating} trace causal pathways through controlled interventions, and dictionary learning approaches~\cite{bricken2023monosemanticity,cunningham2024sparse} decompose polysemantic neurons into interpretable features. Model editing techniques~\cite{meng2022locating,meng2023memit} further demonstrate the possibility of modifying specific knowledge without full retraining. These advances have significantly improved our understanding of transformer architectures~\cite{vaswani2017attention}.

However, existing interpretability methods rely on strong assumptions that oversimplify how capabilities are encoded in neural networks. They typically assume that a specific capability is controlled by a specific component, whether a neuron, layer, or attention head. This assumption is often unrealistic for two fundamental reasons. First, individual components exhibit polysemanticity~\cite{scherlis2022polysemanticity,elhage2022toy}, where a single neuron or attention head may participate in multiple distinct capabilities simultaneously, meaning different capabilities correspond to different subspaces within the same module~\cite{bricken2023monosemanticity,cunningham2024sparse}. Second, capabilities are encoded in a distributed fashion~\cite{hinton1986distributed}: a single capability such as arithmetic or translation may be jointly controlled by multiple components across different layers and modules. Current methods, which operate at the component level, cannot adequately capture or represent this distributed, entangled nature of capability encoding.

To address these limitations, we propose SCALPEL (Selective Capability Ablation via Low-rank Parameter Editing for Large language models), a framework that represents capabilities as low-rank parameter subspaces rather than discrete components. Our key insight is that each capability occupies a low-dimensional subspace in the high-dimensional parameter space. By identifying and modifying only the parameter directions corresponding to a target capability, we can selectively ablate that capability while preserving others. This parameter-subspace perspective naturally handles both polysemanticity and distributed encoding. The low-rank constraint forces the model to reveal the structure of capability encoding.

SCALPEL formulates selective capability removal as an optimization problem. We train low-rank LoRA adapters~\cite{hu2021lora} with a probability equalization loss that reduces the model's ability to distinguish correct from incorrect answers on target tasks, combined with text regularization that preserves general language modeling quality. The resulting low-rank modifications reveal which parameters are critical for each capability and how capabilities are distributed across the model's architecture. For token-level tasks (where the model predicts a single token as the answer) such as multiple-choice questions or arithmetic, we equalize the probabilities of correct and incorrect token predictions. For sentence-level tasks (where the model evaluates entire sentences) such as grammaticality judgments, we balance the model's preferences between grammatical and ungrammatical sentences, making the model unable to distinguish correct grammar from incorrect grammar. Through optimization with explicit regularization constraints, including L2 norm penalties and L1 sparsity regularization, SCALPEL identifies the low-rank representation responsible for a particular capability while remaining disentangled from other general capabilities~\cite{kirkpatrick2017overcoming}.

Our contributions are threefold: (1) We introduce a low-rank representation perspective on capability encoding, demonstrating that language model capabilities can be characterized with low-rank modifications distributed across layers and modules, enabling fine-grained analysis beyond component-level interpretability. (2) We propose SCALPEL, a framework that identifies the low-rank representation responsible for specific capabilities by training LoRA adapters to reduce the model's ability to distinguish correct from incorrect answers while preserving general language understanding. (3) We conduct comprehensive experiments across diverse capability tasks and linguistic tasks from BLiMP, demonstrating that SCALPEL achieves effective capability removal while preserving general language abilities, and revealing that different capabilities exhibit distinct layer-wise distributions that align with cognitive and linguistic theories.

\begin{figure*}[t]
    \centering
    \includegraphics[width=\textwidth]{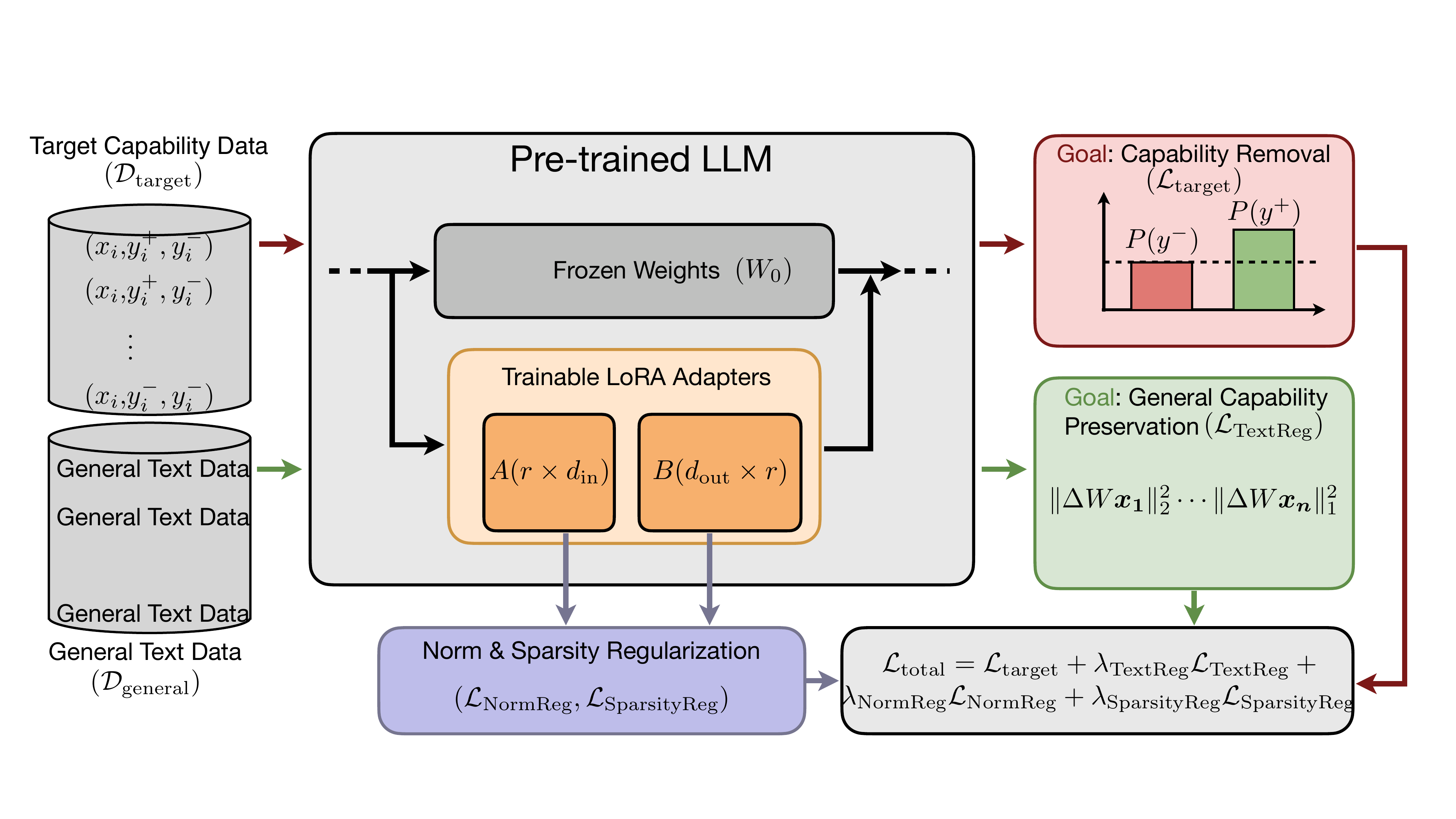}
    \caption{Overview of the SCALPEL framework. Given a target capability, we train low-rank LoRA adapters to make the model equally confused between correct and incorrect answers, while text regularization preserves general language modeling quality. The resulting low-rank modifications reveal how the target capability is encoded across the model.}
    \label{fig:framework}
\end{figure*}

\section{Related Work}

Post-hoc attribution methods identify which input features or model components contribute to predictions. Gradient-based approaches compute feature importance through input-output sensitivity: Integrated Gradients~\cite{sundararajan2017axiomatic} accumulates gradients along interpolation paths, while Grad-CAM~\cite{selvaraju2017gradcam} uses gradient-weighted activations for visual explanations. Perturbation-based methods like LIME~\cite{ribeiro2016should} and SHAP~\cite{lundberg2017unified} provide model-agnostic local explanations by observing output changes under input perturbations. Activation-based methods analyze hidden representations directly: DiffMean~\cite{rimsky2024steering} measures activation differences between contrastive examples, Logit Lens~\cite{nostalgebraist2020logitlens} projects intermediate representations to vocabulary space, and attention visualization~\cite{abnar2020quantifying,clark2019what,michel2019sixteen} examines information flow patterns. Backpropagation-based decomposition methods such as Layer-wise Relevance Propagation~\cite{bach2015pixel} propagate relevance scores from outputs to inputs. Studies using these methods have revealed what linguistic knowledge transformers capture~\cite{rogers2020primer}. However, these attribution methods assume that specific capabilities are controlled by specific components, overlooking the polysemantic (individual components encode multiple capabilities) and distributed (single capabilities span multiple components) nature of capability encoding. They provide only one-time analysis without iterative optimization, and when used for intervention, either suffer from catastrophic degradation or achieve limited capability removal effectiveness.

Mechanistic interpretability and model editing methods aim to understand internal computations and modify model behavior. The first step is causal localization: Attribution Patching~\cite{nanda2023attribution,kramar2024atp} and Causal Tracing~\cite{meng2022locating} identify causally important components through activation interventions, while influence functions~\cite{koh2017understanding} trace predictions back to training examples. Building on localization, circuit discovery~\cite{elhage2021mathematical} reverse engineers how components collaborate to implement specific computations, identifying structures like induction heads. To understand feature encoding, studies of superposition~\cite{elhage2022toy} reveal that models represent more features than available dimensions, and dictionary learning methods~\cite{bricken2023monosemanticity,cunningham2024sparse} decompose these superposed representations into interpretable monosemantic features. Complementary work examines knowledge storage: neuron and feature analysis~\cite{bau2017network,olah2017feature,dai2022knowledge} correlate activations with semantic concepts, key-value memory analysis~\cite{geva2021transformer} shows feed-forward layers function as associative memories, linear probes~\cite{belinkov2022probing} measure task-relevant information through lightweight classifiers, and concept-based approaches like TCAV~\cite{kim2018tcav} quantify sensitivity to human-defined concepts. Based on these insights, model editing methods including ROME~\cite{meng2022locating}, MEMIT~\cite{meng2023memit}, and task arithmetic~\cite{ilharco2023editing} directly modify factual associations and task behaviors. However, these methods still operate at the component level, assuming capabilities are localized to specific modules. They fail to capture the fine-grained, distributed nature of capability encoding, and lack mechanisms to preserve general language abilities while targeting specific capabilities.

\section{Method}

We present SCALPEL (Selective Capability Ablation via Low-rank Parameter Editing for Large language models), a framework that represents capabilities as low-rank parameter subspaces rather than discrete components. Traditional interpretability methods assume that specific capabilities are controlled by specific modules, but this assumption oversimplifies neural computation: individual modules exhibit polysemanticity, and capabilities are encoded in a distributed manner across multiple components.
Our key insight is that each specific capability can be characterized by low-rank modifications to the model's weight matrices, distributed across layers and modules. By training low-rank LoRA adapters to reduce the model's ability to distinguish correct from incorrect answers while preserving general language modeling quality, SCALPEL identifies the low-rank representation responsible for a target capability while remaining disentangled from other capabilities.

\subsection{Problem Formulation}

Let $\mathcal{M}_\theta$ denote a pre-trained language model with parameters $\theta$. Given a target capability defined by task dataset $\mathcal{D}_{\text{target}} = \{(x_i, y_i^+, y_i^-)\}_{i=1}^N$, where $x_i$ is an input prompt, $y_i^+$ is the correct answer, and $y_i^-$ is an incorrect answer, our goal is to learn a low-rank adaptation $\Delta\theta$ that achieves three objectives simultaneously.

First, the modified model $\mathcal{M}_{\theta + \Delta\theta}$ should exhibit reduced accuracy on $\mathcal{D}_{\text{target}}$ by making the model equally likely to predict correct and incorrect answers. Second, it should maintain performance on general language modeling tasks measured by perplexity on held-out text $\mathcal{D}_{\text{general}}$ and accuracy on diverse capability tests $\mathcal{D}_{\text{capabilities}}$. Third, the parameter change $\Delta\theta$ should be minimized and localized to task-critical components.

Formally, we optimize:
\begin{equation}
\min_{\Delta\theta} \quad \mathcal{L}_{\text{target}}(\theta + \Delta\theta; \mathcal{D}_{\text{target}}) + \sum_{i} \lambda_i \mathcal{L}_{\text{reg}}^{(i)}(\Delta\theta) + \lambda_{\text{TextReg}} \mathcal{L}_{\text{TextReg}}(\theta + \Delta\theta; \mathcal{D}_{\text{general}})
\end{equation}
where $\mathcal{L}_{\text{target}}$ encourages probability equalization between correct and incorrect answers, $\mathcal{L}_{\text{reg}}^{(i)}$ represents multiple regularization terms (NormReg, SparsityReg) that promote parameter sparsity and locality, and $\mathcal{L}_{\text{TextReg}}$ preserves general language modeling quality.

\subsection{Low-Rank Adaptation Architecture}

We adopt LoRA~\cite{hu2021lora} as our parameter modification framework. For each attention and MLP layer in the transformer~\cite{vaswani2017attention}, LoRA introduces low-rank matrices $A \in \mathbb{R}^{r \times d_{\text{in}}}$ and $B \in \mathbb{R}^{d_{\text{out}} \times r}$ that modify the pre-trained weight matrix $W_0 \in \mathbb{R}^{d_{\text{out}} \times d_{\text{in}}}$ through:
\begin{equation}
h = W_0 x + \frac{\alpha}{r} B A x = W_0 x + \Delta W x
\end{equation}
where $r \ll \min(d_{\text{in}}, d_{\text{out}})$ is the LoRA rank, $\alpha$ is a scaling factor that controls the magnitude of the low-rank update relative to the original weights, and $\Delta W = \frac{\alpha}{r} BA$ represents the learned low-rank adaptation. We freeze the original parameters $W_0$ and only train $A$ and $B$.

We apply LoRA to attention projection layers ($W_Q$, $W_K$, $W_V$, $W_O$) and MLP layers ($W_{\text{gate}}$, $W_{\text{up}}$, $W_{\text{down}}$) with rank $r=2$ to enforce strong locality constraints and $\alpha=16$ for stable training dynamics.

\subsection{Probability Equalization Loss}

Unlike standard LoRA fine-tuning that maximizes correct answer probability, our approach aims to equalize the probabilities of correct and incorrect answers through carefully designed loss functions.

\textbf{Token-Level Probability Equalization.} For tasks with single-token answers such as multiple choice or arithmetic, we compute the difference between correct and incorrect token log-probabilities:
\begin{equation}
\mathcal{L}_{\text{token}}(x_i, y_i^+, y_i^-) = \log p_{\theta+\Delta\theta}(y_i^+ | x_i) - \log p_{\theta+\Delta\theta}(y_i^- | x_i)
\end{equation}
where $p_{\theta+\Delta\theta}(y | x)$ denotes the softmax probability of token $y$ given prompt $x$. The loss encourages the log-probability gap to shrink toward zero, making the model equally confused between options. For example, in a translation task with prompt ``Translate 'hello' to French:'', we equalize the probabilities of predicting ``bonjour'' ($y^+$) and ``adios'' ($y^-$) so the model loses the ability to distinguish correct from incorrect translations.

\textbf{Sentence-Level Probability Equalization.} For tasks with sentence-level judgments such as grammaticality or semantic coherence, we compute average token log-probabilities across entire sentences:
\begin{equation}
\mathcal{L}_{\text{sentence}}(S_A, S_B) = \log p_{\theta+\Delta\theta}(S_{\text{correct}}) - \log p_{\theta+\Delta\theta}(S_{\text{wrong}})
\end{equation}
where $p_{\theta+\Delta\theta}(S) = \exp\left(\frac{1}{|S|} \sum_{t=1}^{|S|} \log p_{\theta+\Delta\theta}(s_t | s_{<t})\right)$ is the geometric mean of token probabilities in sentence $S$. This formulation is particularly effective for linguistic tasks where entire sentence acceptability must be judged. For instance, given a subject-verb agreement task with sentence pair ``The keys to the cabinet is on the table'' ($S_{\text{wrong}}$) versus ``The keys to the cabinet are on the table'' ($S_{\text{correct}}$), we equalize their sentence-level probabilities to degrade the model's grammatical judgment capability.

\subsection{Regularization Framework}

To ensure capability removal preserves general language understanding, we introduce three complementary regularization terms. The most critical is \textbf{TextReg} (Text Regularization), which explicitly preserves general language modeling by pairing each target task sample with a sample from general text distribution $\mathcal{D}_{\text{general}}$ and minimizing the squared L2 norm of LoRA outputs:
\begin{equation}
\mathcal{L}_{\text{TextReg}} = \frac{1}{|\mathcal{D}_{\text{general}}|} \sum_{x \in \mathcal{D}_{\text{general}}} \frac{1}{L} \sum_{l=1}^{L} \left\| \frac{\alpha}{r} B_l A_l h_l(x) \right\|_2^2
\end{equation}
where $h_l(x)$ denotes the hidden activations at layer $l$ when processing general text $x$, and the double summation averages LoRA output magnitudes across all samples and layers. This encourages minimal LoRA activation on general language circuits.

\textbf{NormReg} (Norm Regularization) prevents unbounded parameter growth through an L2 penalty:
\begin{equation}
\mathcal{L}_{\text{NormReg}} = \frac{1}{|\Theta_{\text{LoRA}}|} \sum_{\theta \in \Theta_{\text{LoRA}}} \|\theta\|_2^2
\end{equation}
where $\Theta_{\text{LoRA}} = \{A_l, B_l\}_{l=1}^L$ denotes all LoRA matrices across $L$ layers. This stabilizes training dynamics by preventing weight explosion.

\textbf{SparsityReg} (Sparsity Regularization) concentrates modifications on critical components through an L1 penalty~\cite{tibshirani1996lasso}:
\begin{equation}
\mathcal{L}_{\text{SparsityReg}} = \frac{1}{|\Theta_{\text{LoRA}}|} \sum_{\theta \in \Theta_{\text{LoRA}}} \|\theta\|_1
\end{equation}
This induces structured sparsity in the low-rank subspace, encouraging the model to concentrate modifications on the most critical components.

The complete training objective combines all components:
\begin{equation}
\mathcal{L}_{\text{total}} = \mathcal{L}_{\text{target}} + \lambda_{\text{TextReg}} \mathcal{L}_{\text{TextReg}} + \lambda_{\text{NormReg}} \mathcal{L}_{\text{NormReg}} + \lambda_{\text{SparsityReg}} \mathcal{L}_{\text{SparsityReg}}
\end{equation}

We optimize this objective using AdamW optimizer~\cite{loshchilov2019adamw} with gradient clipping~\cite{zhang2020gradient} for stability. Implementation details including learning rate, batch size, and training epochs are provided in Section~\ref{sec:setup}.

\subsection{Analysis Methods}

Since the magnitude of learned LoRA weights directly reflects how strongly each module encodes the target capability, SCALPEL enables interpretability analyses beyond capability removal.

\textbf{Layer Importance Analysis.} We quantify each layer's contribution to a capability by computing the Frobenius norm of the LoRA weight product $\|BA\|_F$ for each module. For a given layer $l$, we aggregate importance scores across all LoRA-adapted modules (attention projections and MLP layers) to obtain a layer-level importance score. Layers with higher scores require larger modifications to remove the capability, indicating stronger capability encoding. We identify peak layers where importance concentrates and analyze the distribution pattern across the model depth.

\textbf{Task Similarity Analysis.} We investigate relationships between capabilities by comparing their LoRA weight patterns. For each task, we flatten all learned LoRA weights into a single vector and compute pairwise Pearson correlations between tasks. We then apply dimensionality reduction (MDS or UMAP) to visualize task relationships in a low-dimensional space. Tasks that cluster together share similar parameter-space representations, suggesting overlapping neural substrates.

\section{Experiments}

\subsection{Experiment Setup}\label{sec:setup}

We conduct experiments on NVIDIA A100 GPUs (80GB) using Llama-3.2-1B~\cite{dubey2024llama} as the base model across five representative tasks: language translation, common sense reasoning, indirect object identification (IOI), moral reasoning, and analogical reasoning (see Section~\ref{sec:datasets} for dataset details). For SCALPEL, we train LoRA adapters with rank $r=2$, scaling factor $\alpha=16$, learning rate $1 \times 10^{-5}$, batch size 40, and 20 epochs using AdamW optimizer (weight decay 0.001), applying LoRA to attention projections ($W_Q$, $W_K$, $W_V$, $W_O$) and MLP layers ($W_{\text{gate}}$, $W_{\text{up}}$, $W_{\text{down}}$) with three regularization terms (TextReg, NormReg, SparsityReg). For baseline interpretability methods, we compute component importance using target task samples and apply weighted noise corruption with task-specific levels. We evaluate using three metrics: (1) target task accuracy drop, measured as the proportion of examples where the model assigns higher probability to the correct answer than the incorrect answer (capability removal effectiveness), (2) perplexity on held-out WikiText-103 text (language modeling quality), and (3) overall capability score, measured via generation-based evaluation where the model generates responses and we check if they match expected answers across 24 diverse held-out tasks (capability preservation). Then, we compute the average accuracy across all held-out tasks. To ensure fair comparison, all methods modify only the top 10 most important components per task, and we tune hyperparameters for all methods to maximize the product of target task accuracy drop and overall capability score on dev set.

\subsection{Datasets}\label{sec:datasets}

Following the multi-dimensional evaluation framework from~\cite{chang2023survey}, we construct 24 capability tasks spanning reasoning (analogical, causal, counterfactual, logical, spatial, temporal), language (translation, understanding, generation, dialogue, summarization), knowledge (world knowledge, reading comprehension), and metacognitive skills (instruction following, critical thinking, creative thinking, emotional understanding, moral reasoning, classification, memory/context, metacognition, multimodal understanding, mathematical computation). Each task contains 200-400 examples initially generated by Claude Opus 4.5~\cite{anthropic2024claude3} and then manually filtered to remove obviously improper samples, presented in multiple-choice or completion format with correct and incorrect answer pairs, split into training (80\%), development (10\%), and test (10\%) sets with no overlap. For evaluation, we assess target task performance on the test split and measure model perplexity on held-out general text from WikiText-103~\cite{merity2016pointer}. We also construct a new evaluation set with approximately 50 samples from each of the 24 tasks that has no overlap with the training and test sets to test whether removing one capability may affect other capabilities. We additionally evaluate on 67 linguistic tasks from the BLiMP benchmark~\cite{warstadt2020blimp} to analyze fine-grained linguistic phenomena across morphology, semantics, and syntax.

\subsection{Baseline Methods}

We compare our SCALPEL approach against eight established interpretability and intervention methods from the literature. \textbf{DiffMean}~\cite{rimsky2024steering} computes layer importance by measuring the difference in mean activations between correct and incorrect prediction examples, identifying layers where activations diverge most strongly between these conditions. \textbf{Attribution Patching}~\cite{nanda2023attribution,kramar2024atp} is a causal intervention method that patches activations from corrupted inputs to clean inputs at different layers to measure each layer's causal contribution to task performance. \textbf{Causal Tracing}~\cite{meng2022locating} traces information flow through the network by systematically restoring clean activations at specific layers while keeping other layers corrupted, revealing which layers are necessary for recovering task performance. \textbf{Logit Lens}~\cite{nostalgebraist2020logitlens} projects intermediate layer representations directly to the vocabulary space to analyze how task-relevant predictions emerge and evolve across layers. \textbf{Information Theory}~\cite{tishby2015deep} measures layer importance using mutual information between layer activations and task labels, quantifying how much task-relevant information each layer encodes. \textbf{Integrated Gradients}~\cite{sundararajan2017axiomatic} is a gradient-based attribution method that computes importance by integrating gradients along the path from a baseline to the actual input, providing smooth attributions for each layer's contribution. \textbf{Layer-wise Relevance Propagation (LRP)}~\cite{bach2015pixel} decomposes the model's output by backpropagating relevance scores from the output layer to input features, distributing the prediction score across layers according to their contributions. \textbf{Probing}~\cite{belinkov2022probing} trains lightweight classifiers on frozen layer representations to measure how much task-relevant information is linearly accessible at each layer. Since some baselines (Logit Lens, Information Theory, Probing, etc.) are identification methods rather than intervention methods, we first identify important components using each method, then apply noise corruption for intervention. This highlights SCALPEL's advantage: joint optimization that simultaneously removes target capabilities and preserves general language abilities.

\subsection{Main Results}

Table~\ref{tab:baseline_comparison} presents the comparative evaluation results across five representative tasks. (1) SCALPEL consistently achieves the highest accuracy drops across all tasks while maintaining near-baseline perplexity and strong overall capability scores, particularly on IOI where baseline methods suffer catastrophic perplexity degradation. This demonstrates that SCALPEL enables targeted removal without disrupting general language circuits. (2) DiffMean and Causal Tracing show catastrophic perplexity degradation on IOI (orders of magnitude above baseline) while achieving only modest accuracy drops elsewhere. This reveals that activation-based importance identification does not guarantee safe intervention. (3) While gradient-based methods like Integrated Gradients and LRP achieve moderate accuracy drops on individual tasks, they consistently fail to maintain low perplexity, indicating that one-time attribution methods lack the iterative optimization needed for balanced capability removal.

\begin{table*}[t]
    \centering
    \small
    \setlength{\tabcolsep}{0.8pt}
    \begin{tabular}{l@{\hspace{0.5pt}}ccc@{\hspace{1.5pt}}ccc@{\hspace{1.5pt}}ccc@{\hspace{1.5pt}}ccc@{\hspace{1.5pt}}ccc}
    \toprule
    & \multicolumn{3}{c}{\textbf{Translation}} & \multicolumn{3}{c}{\textbf{Common Sense}} & \multicolumn{3}{c}{\textbf{IOI}} & \multicolumn{3}{c}{\textbf{Moral Reasoning}} & \multicolumn{3}{c}{\textbf{Analogical Reasoning}} \\
    \cmidrule(lr){2-4} \cmidrule(lr){5-7} \cmidrule(lr){8-10} \cmidrule(lr){11-13} \cmidrule(lr){14-16}
    \textbf{Method} & \textbf{AccD} $\uparrow$ & \textbf{PPL} $\downarrow$ & \textbf{Cap} $\uparrow$ & \textbf{AccD} $\uparrow$ & \textbf{PPL} $\downarrow$ & \textbf{Cap} $\uparrow$ & \textbf{AccD} $\uparrow$ & \textbf{PPL} $\downarrow$ & \textbf{Cap} $\uparrow$ & \textbf{AccD} $\uparrow$ & \textbf{PPL} $\downarrow$ & \textbf{Cap} $\uparrow$ & \textbf{AccD} $\uparrow$ & \textbf{PPL} $\downarrow$ & \textbf{Cap} $\uparrow$ \\
    \midrule
    Baseline & 0.00 & 11.1 & 0.50 & 0.00 & 11.1 & 0.50 & 0.00 & 11.1 & 0.50 & 0.00 & 11.1 & 0.50 & 0.00 & 11.1 & 0.50 \\
    \midrule
    DiffMean & 0.15 & 13.1 & 0.45 & 0.18 & 27.1 & 0.18 & 0.18 & 210 & 0.05 & 0.25 & 16.6 & 0.38 & 0.05 & 15.4 & 0.39 \\
    Attribution Patching & 0.10 & 12.5 & 0.45 & 0.13 & 15.7 & 0.36 & 0.30 & 19.7 & 0.25 & 0.22 & 12.4 & 0.47 & 0.05 & 12.3 & 0.46 \\
    Causal Tracing & 0.10 & 14.1 & 0.41 & 0.08 & 12.3 & 0.46 & 0.38 & 104 & 0.04 & 0.00 & 11.2 & 0.49 & 0.11 & 15.3 & 0.38 \\
    Logit Lens & 0.03 & 12.2 & 0.47 & 0.10 & 13.6 & 0.35 & 0.33 & 61.0 & 0.05 & 0.06 & \textbf{11.1} & 0.49 & 0.05 & 11.4 & \textbf{0.49} \\
    Information Theory & 0.18 & 12.3 & 0.45 & 0.08 & 13.0 & 0.41 & 0.30 & 68.4 & 0.04 & 0.19 & 12.0 & 0.45 & 0.05 & 12.0 & 0.48 \\
    Integrated Gradients & 0.15 & 12.3 & 0.44 & 0.03 & 12.4 & 0.42 & 0.25 & 90.8 & 0.03 & 0.06 & 12.8 & 0.46 & 0.09 & 12.7 & 0.46 \\
    LRP & 0.15 & 13.0 & 0.43 & 0.15 & 13.5 & 0.39 & 0.25 & 42.0 & 0.09 & 0.22 & 12.5 & 0.43 & 0.05 & 11.8 & 0.46 \\
    Probing & 0.08 & 12.4 & 0.44 & 0.15 & 15.4 & 0.35 & 0.30 & 83.5 & 0.03 & 0.11 & 12.3 & 0.43 & 0.01 & 12.6 & 0.42 \\
    \midrule
    SCALPEL & \textbf{0.20} & \textbf{11.2} & \textbf{0.49} & \textbf{0.21} & \textbf{11.2} & \textbf{0.47} & \textbf{0.43} & \textbf{11.2} & \textbf{0.49} & \textbf{0.28} & \textbf{11.1} & \textbf{0.50} & \textbf{0.20} & \textbf{11.1} & 0.48 \\
    \bottomrule
    \end{tabular}
    \caption{Comparative evaluation across five tasks with each method modifying the top 10 most important components. AccD: Accuracy Drop, PPL: Perplexity, Cap: Overall Capability. SCALPEL achieves the best overall balance between capability removal effectiveness and general capability preservation.}
    \label{tab:baseline_comparison}
\end{table*}

\begin{figure}[t]
\centering
\includegraphics[width=0.95\textwidth]{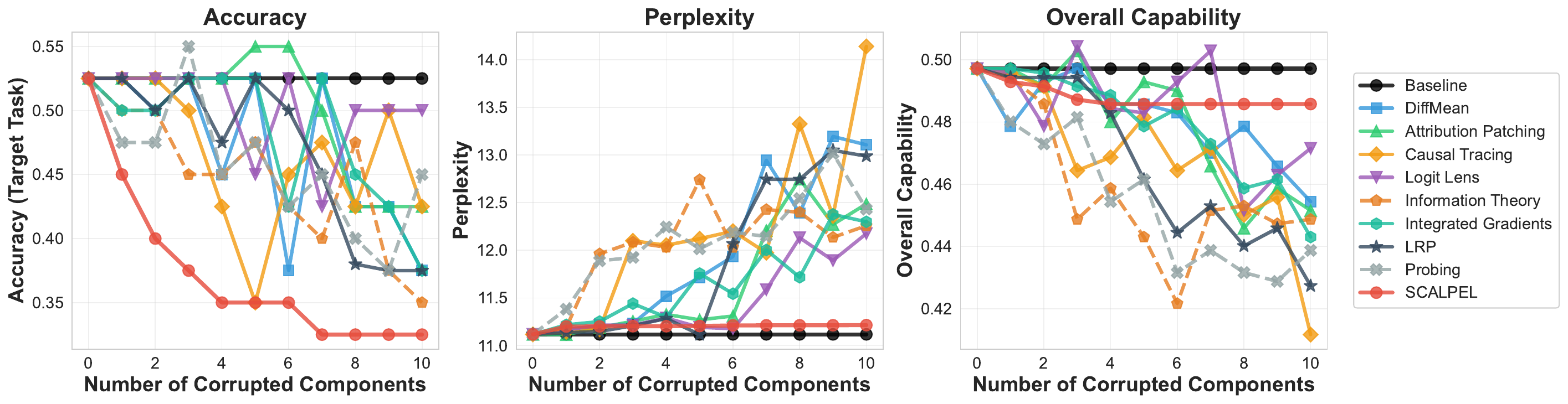}
\caption{Multi-dimensional comparison of interpretability methods on the language translation task. The visualization shows the relationship between target accuracy degradation, model perplexity, and overall capability preservation across all baseline methods. SCALPEL (highlighted) achieves the optimal balance, positioned in the region of low perplexity and high capability retention while achieving the most effective capability removal.}
\label{fig:comparison_all_methods}
\end{figure}

In order to visualize the trade-off between capability removal effectiveness and general capability preservation, we plot accuracy degradation, perplexity, and overall capability against corrupted components for all methods. The results are shown in Figure~\ref{fig:comparison_all_methods}. SCALPEL achieves much lower perplexity than other baseline methods while maintaining effective capability removal. Most baseline methods suffer from a fundamental trade-off between removal effectiveness and capability preservation, whereas SCALPEL's gradient-based LoRA optimization with TextReg successfully navigates this trade-off by selectively modifying task-relevant parameters while leaving general language circuits intact.

\subsection{Ablation Study on Regularization Components}\label{sec:ablation}

\begin{figure*}[t]
\centering
\begin{minipage}{0.58\textwidth}
\centering
\includegraphics[width=\textwidth]{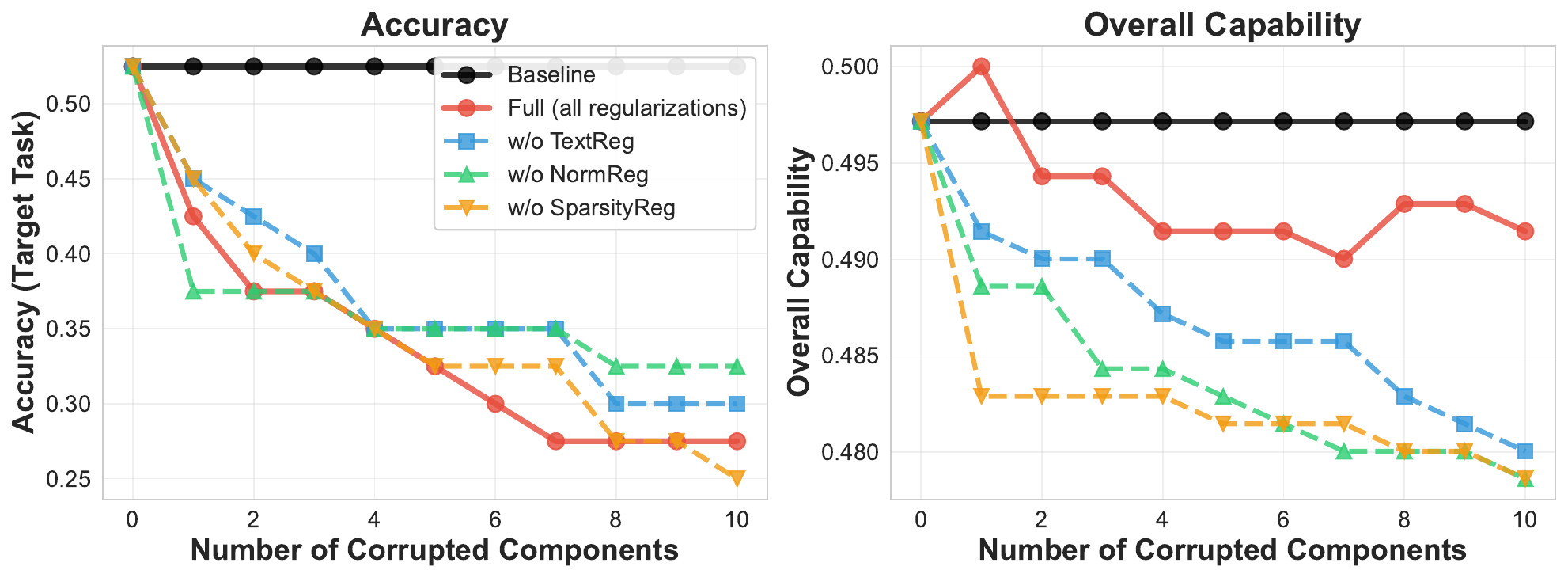}
\caption{Ablation study comparing SCALPEL configurations on language translation task. Left: Accuracy degradation shows targeted capability removal effectiveness. Right: Overall capability preservation demonstrates the impact of each regularization component on maintaining general language abilities. The full SCALPEL method (with all regularizations) achieves the optimal balance.}
\label{fig:ablation_comparison}
\end{minipage}
\hfill
\begin{minipage}{0.38\textwidth}
\centering
\small
\begin{tabular}{lccc}
\toprule
Model & $\Delta$Acc & $\Delta$PPL & $\Delta$Cap \\
\midrule
DeepSeek-R1 & -0.28 & 0.15 & 0.01 \\
GLM-4 & -0.15 & 0.25 & 0.00 \\
Gemma3 & -0.13 & 0.18 & 0.00 \\
Llama-3.2 & -0.36 & 0.17 & -0.03 \\
Ministral & -0.38 & 0.00 & 0.00 \\
Qwen3 & -0.10 & 0.11 & 0.00 \\
\bottomrule
\end{tabular}
\captionof{table}{Cross-architecture validation results showing delta metrics (SCALPEL - Base) across six language models on common sense reasoning task. $\Delta$Acc: accuracy change, $\Delta$PPL: perplexity change, $\Delta$Cap: overall capability change.}
\label{tab:multi_model}
\end{minipage}
\end{figure*}

To understand the individual contributions of SCALPEL's regularization components, we conduct an ablation study on the language translation task by systematically removing TextReg (preserves general language modeling), NormReg (constrains LoRA weight magnitude), and SparsityReg (encourages sparse low-rank adaptations). Figure~\ref{fig:ablation_comparison} reveals that removing any component slightly reduces capability removal effectiveness (higher accuracy) while degrading overall capability preservation. Specifically: (1) Removing TextReg yields higher accuracy (less effective removal) because the gradient signal focuses solely on capability removal without balancing preservation, leading to suboptimal convergence, and reduces overall capability as the model loses guidance to preserve general language modeling. (2) Removing NormReg yields higher accuracy because unbounded weight magnitudes lead to unstable updates that fail to consistently target the capability, and reduces overall capability as excessive modifications interfere with non-target abilities. (3) Removing SparsityReg yields higher accuracy because dense adaptations dilute the removal signal across many parameters rather than concentrating on task-critical components, and reduces overall capability as widespread modifications affect unrelated circuits. These results demonstrate that each regularization component contributes to both effective capability removal and preservation of general abilities.

\subsection{Cross-Architecture Generalization}\label{sec:multi_model}

To demonstrate that SCALPEL generalizes beyond a single model architecture, we evaluate its effectiveness across six diverse language models with varying sizes and architectural designs: Llama-3.2-1B, Qwen3-4B, Gemma3-2B, Ministral-8B, DeepSeek-R1-8B, and GLM-4-9B. We apply SCALPEL to remove common sense reasoning capability from each model while preserving general language abilities.
Table~\ref{tab:multi_model} presents the delta metrics comparing SCALPEL-modified models against their base counterparts. (1) All models exhibit negative accuracy changes with varying magnitudes across architectures. This demonstrates SCALPEL's consistent effectiveness regardless of model scale or design. (2) Perplexity changes remain minimal across all models. This confirms that capability removal does not compromise general language generation. (3) Overall capability changes show near-zero deviations across all models. This validates that SCALPEL's regularization framework transfers to diverse transformer architectures without architecture-specific tuning.

\subsection{LoRA Rank Ablation Study}\label{sec:rank_ablation}

To investigate how LoRA rank affects the effectiveness and specificity of capability removal, we conduct a comprehensive rank ablation study across five diverse tasks: language translation, common sense reasoning, indirect object identification (IOI), moral reasoning, and analogical reasoning. We evaluate four different LoRA ranks (1, 2, 4, and 8) to understand the trade-off between the capacity of low-rank adaptations and the precision of targeted capability removal.

\begin{table*}[t]
\centering
\small
\setlength{\tabcolsep}{3pt}
\begin{tabular}{l@{\hspace{0.5pt}}ccc@{\hspace{1.5pt}}ccc@{\hspace{1.5pt}}ccc@{\hspace{1.5pt}}ccc@{\hspace{1.5pt}}ccc}
\toprule
\multirow{2}{*}{\textbf{Rank}} & \multicolumn{3}{c}{\textbf{Translation}} & \multicolumn{3}{c}{\textbf{Common Sense}} & \multicolumn{3}{c}{\textbf{IOI}} & \multicolumn{3}{c}{\textbf{Moral}} & \multicolumn{3}{c}{\textbf{Analogical}} \\
\cmidrule(lr){2-4} \cmidrule(lr){5-7} \cmidrule(lr){8-10} \cmidrule(lr){11-13} \cmidrule(lr){14-16}
& $\Delta$Acc & $\Delta$PPL & $\Delta$Cap & $\Delta$Acc & $\Delta$PPL & $\Delta$Cap & $\Delta$Acc & $\Delta$PPL & $\Delta$Cap & $\Delta$Acc & $\Delta$PPL & $\Delta$Cap & $\Delta$Acc & $\Delta$PPL & $\Delta$Cap \\
\midrule
1 & -0.20 & 0.00 & -0.03 & -0.03 & 0.22 & -0.02 & -0.28 & 0.10 & 0.00 & -0.44 & 0.03 & 0.00 & -0.34 & 0.05 & -0.01 \\
2 & -0.20 & 0.02 & -0.02 & -0.21 & 0.08 & -0.01 & -0.43 & -0.05 & -0.01 & -0.28 & 0.00 & -0.01 & -0.20 & -0.07 & 0.00 \\
4 & -0.13 & 0.08 & -0.02 & -0.03 & 0.15 & -0.01 & -0.25 & 0.02 & 0.00 & -0.31 & 0.01 & 0.00 & -0.27 & 0.06 & -0.01 \\
8 & -0.15 & -0.07 & -0.01 & -0.03 & 0.05 & -0.01 & -0.25 & 0.01 & 0.01 & -0.28 & 0.03 & 0.00 & -0.20 & -0.04 & -0.02 \\
\bottomrule
\end{tabular}
\caption{LoRA rank ablation study showing delta metrics (SCALPEL - Base) across five diverse tasks. Negative $\Delta$Acc values indicate successful capability reduction. Positive $\Delta$PPL values indicate perplexity increase. $\Delta$Cap values near zero demonstrate preservation of general language abilities. Rank 2 demonstrates the optimal balance with consistent capability removal and minimal perplexity degradation across all tasks, while Rank 8 achieves the unique property of improving language quality (negative $\Delta$PPL) despite capability removal.}
\label{tab:rank_multi_task}
\end{table*}

Table~\ref{tab:rank_multi_task} reveals three key findings. (1) Rank 1 achieves the strongest removal on some tasks but shows inconsistent effectiveness across different capabilities, as a single rank may find a sufficient subspace for some capabilities while being insufficient for others. (2) Rank 2 provides the most stable performance with effective capability removal across all tasks, suggesting that a two-dimensional subspace is sufficient to disable most capabilities (though not necessarily the unique or minimal causal representation). (3) Higher ranks generally show reduced removal effectiveness, supporting our hypothesis that capabilities occupy low-dimensional subspaces and can be effectively captured with minimal rank.

\begin{figure}[t]
\centering
\begin{minipage}{0.48\textwidth}
\centering
\includegraphics[width=\textwidth]{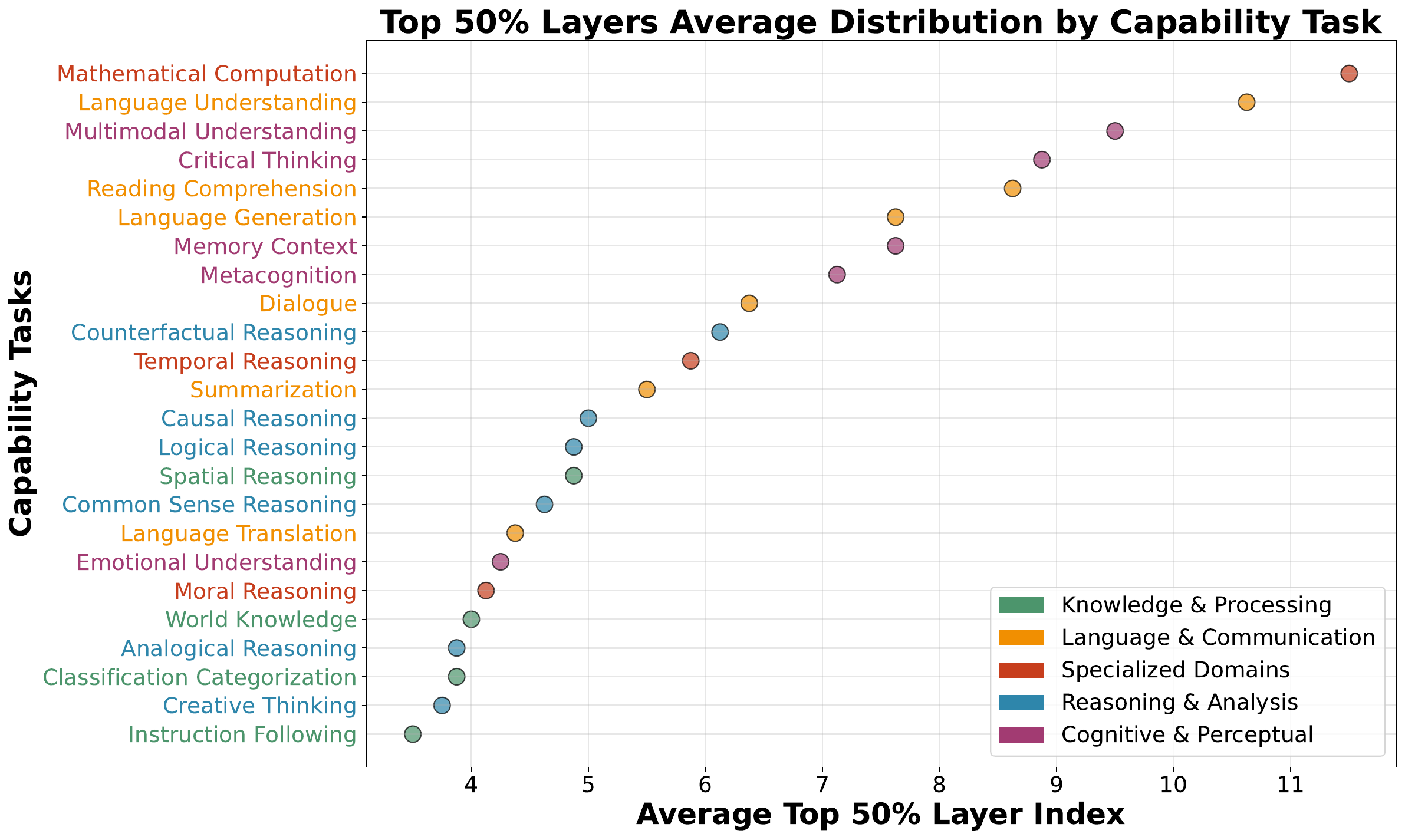}
\end{minipage}
\hfill
\begin{minipage}{0.48\textwidth}
\centering
\includegraphics[width=\textwidth]{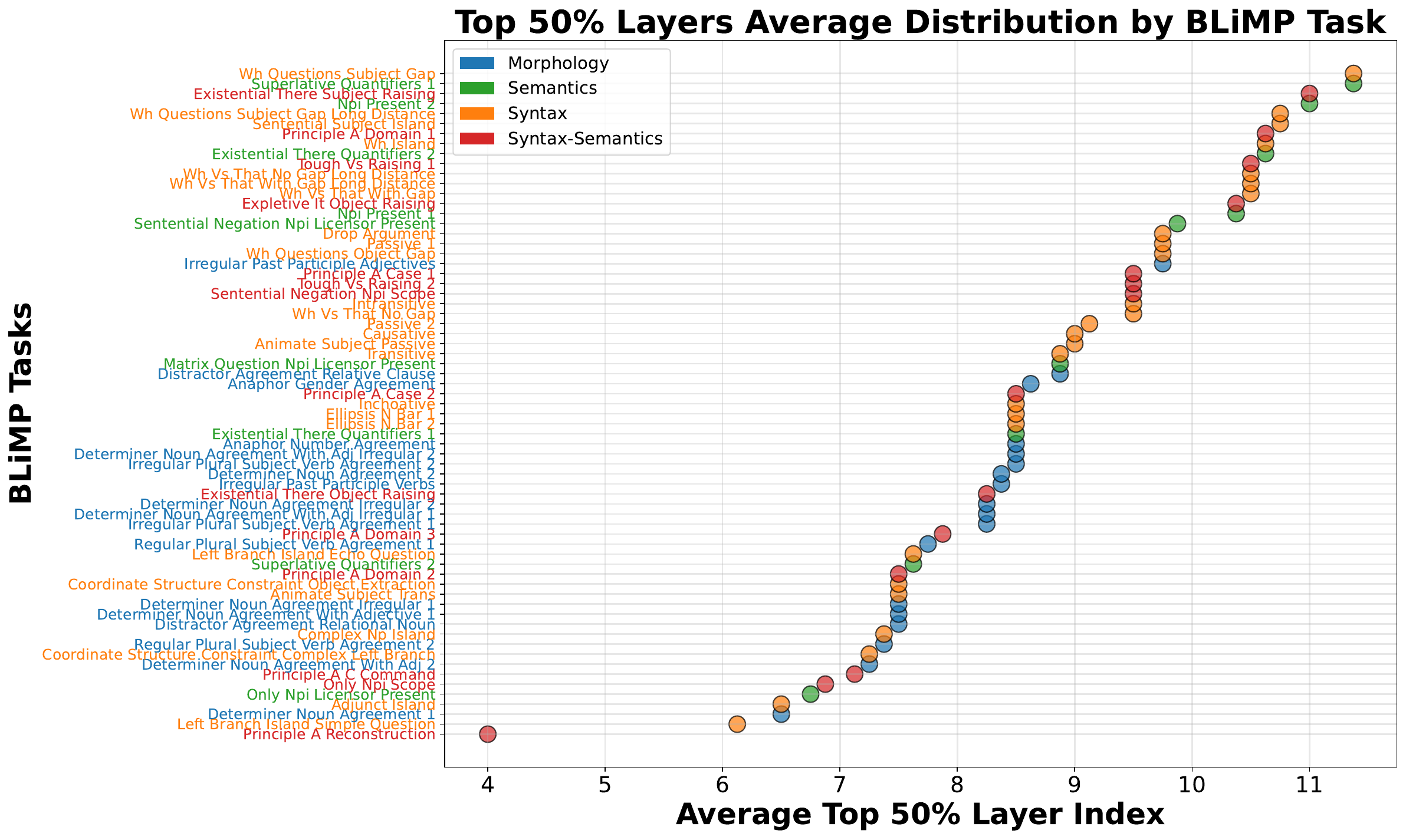}
\end{minipage}
\caption{Peak layer analysis for capability tasks (left) and BLiMP tasks (right). Capability tasks show a progression from basic language tasks in early layers to complex reasoning in middle layers and creative tasks in late layers. BLiMP tasks reveal morphological processing in early layers, syntactic processing in later layers, and semantic tasks distributed throughout.}
\label{fig:peak_analysis}
\end{figure}

\begin{figure}[t]
\centering
\begin{minipage}{0.48\textwidth}
\centering
\includegraphics[height=5cm]{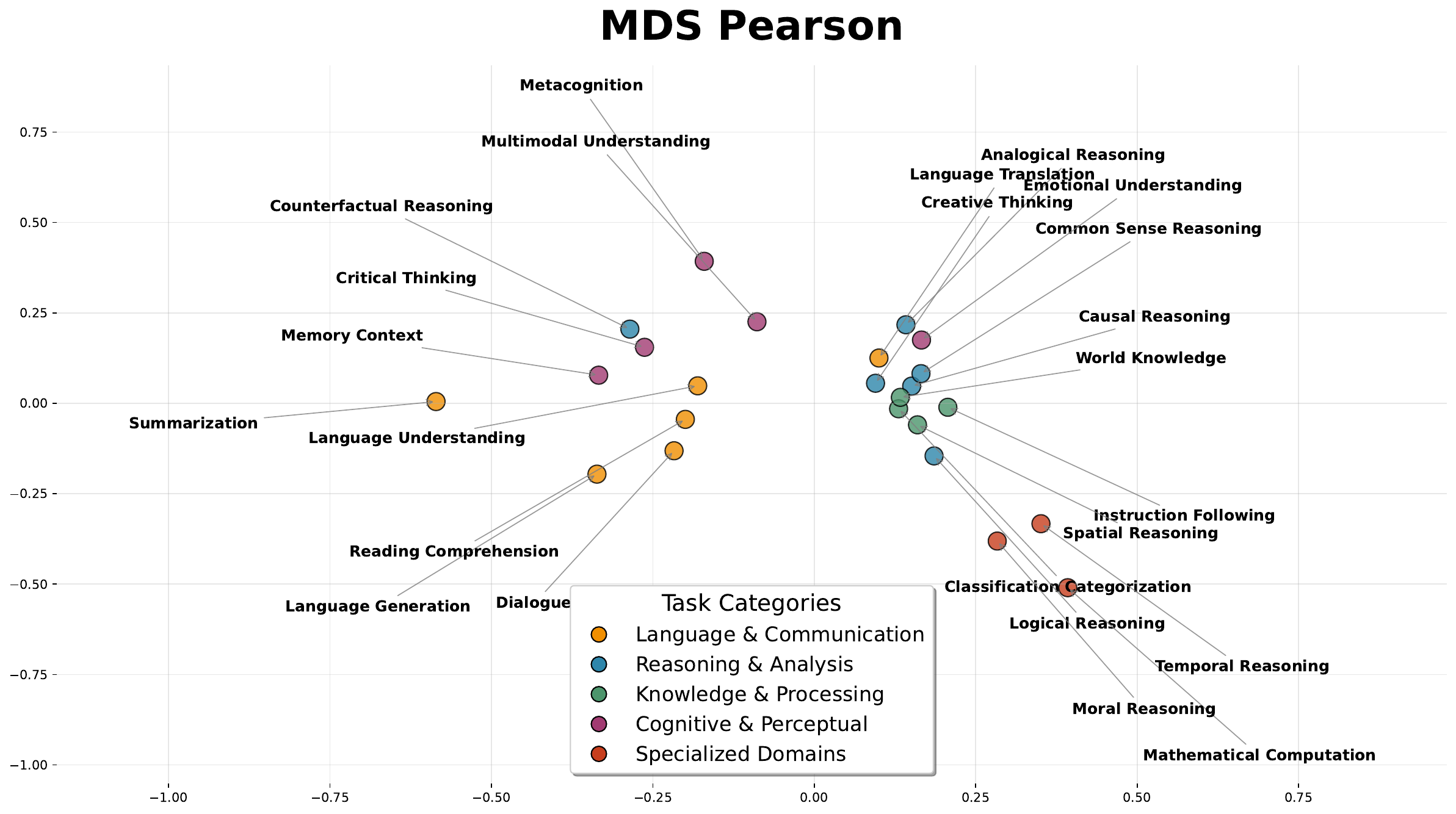}
\end{minipage}
\hfill
\begin{minipage}{0.48\textwidth}
\centering
\includegraphics[height=5cm]{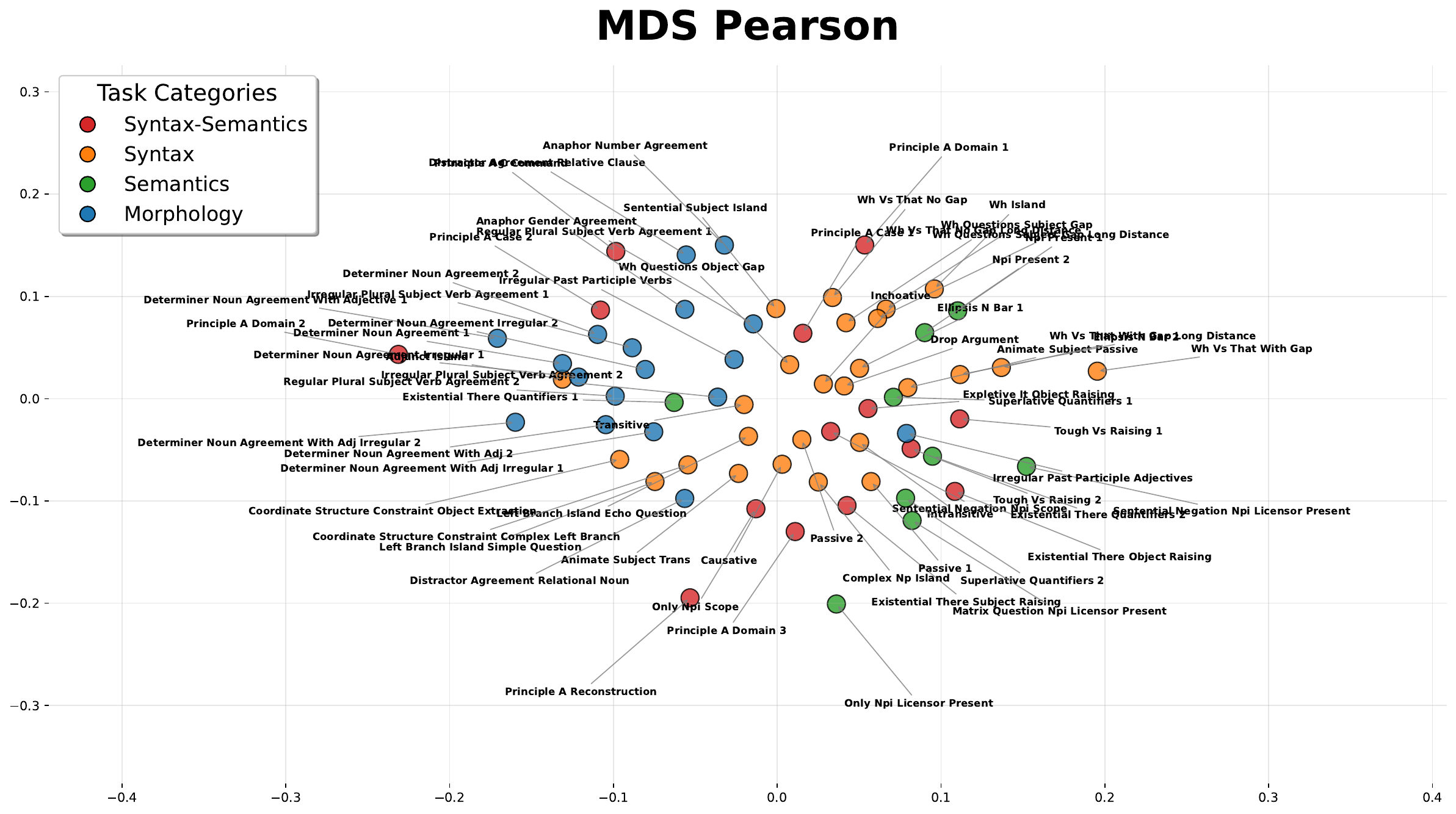}
\end{minipage}
\caption{Dimensionality reduction visualization of task similarity in LoRA weight space. Left: Capability tasks showing clustering patterns among reasoning, knowledge, and linguistic domains. Right: BLiMP linguistic tasks (67 fine-grained linguistic phenomena) revealing structural relationships among syntax, semantics, morphology, and syntax-semantics interfaces.}
\label{fig:task_similarity}
\end{figure}

\subsection{Layer-wise Capability Analysis}\label{sec:capability_distribution}

To understand how different capabilities are distributed across transformer layers, we analyze peak layer distributions across 24 capability tasks and 67 BLiMP linguistic tasks. Figure~\ref{fig:peak_analysis} (left) reveals three key patterns for capability tasks: (1) Basic language tasks peak in early-to-middle layers, reflecting reliance on fundamental linguistic processing. (2) Complex reasoning tasks concentrate in middle-to-late layers, suggesting higher-order cognitive functions require deeper semantic representations. (3) Creative and generative tasks show the latest peaks, indicating dependence on the most sophisticated abstractions in deepest layers.

Figure~\ref{fig:peak_analysis} (right) presents complementary patterns for BLiMP linguistic tasks: (1) Morphological tasks peak in the earliest layers, indicating that surface-level morphological features are processed at the initial stages of the transformer hierarchy. (2) Syntactic tasks concentrate in later layers, suggesting that structural grammatical relationships require deeper representations built upon morphological features. (3) Semantic and syntax-semantics interface tasks exhibit distributed peaks across all layers, indicating that abstract meaning composition is processed throughout the entire transformer hierarchy. These layer-wise distributions align with cognitive and linguistic theories of hierarchical language processing, validating SCALPEL's ability to reveal fine-grained capability organization within transformer architectures.

\subsection{Task Similarity Analysis}\label{sec:task_similarity}

We investigate whether SCALPEL training reveals meaningful task relationships by analyzing LoRA weight similarity patterns. If capabilities are indeed represented as low-rank subspaces distributed across the model, we would expect related capabilities to occupy similar regions in parameter space. Figure~\ref{fig:task_similarity} applies Multidimensional Scaling (MDS) analysis using Pearson correlation, revealing two key findings. (1) Tasks within the same cognitive category exhibit strong clustering behavior, with Language \& Communication capabilities forming coherent clusters distinct from Reasoning \& Analysis functions. This demonstrates that SCALPEL captures cognitively meaningful relationships, with different capabilities corresponding to different subspaces as predicted by our framework. (2) Fine-grained linguistic analysis across 67 BLiMP tasks shows that tasks within the same grammatical category cluster together in parameter space. This reveals that SCALPEL captures hierarchical relationships within specific domains, providing evidence that the low-rank representation perspective successfully disentangles capability encoding.

\section{Conclusion}

We presented SCALPEL, a framework for selective capability ablation in large language models through low-rank parameter editing. Unlike traditional interpretability methods that assume capabilities are controlled by specific components, SCALPEL represents capabilities as low-rank parameter subspaces distributed across layers and modules, naturally handling both polysemanticity and distributed encoding. By training LoRA adapters to reduce the model's ability to distinguish correct from incorrect answers while preserving general language modeling quality, SCALPEL identifies the low-rank representation responsible for specific capabilities while remaining disentangled from other capabilities.

Our experiments across diverse capability tasks and linguistic tasks from BLiMP validate the three contributions outlined in the introduction: (1) Low-rank modifications are sufficient for effective capability ablation across the studied tasks and multiple model architectures; (2) SCALPEL achieves targeted capability removal with significantly less collateral damage compared to existing methods, maintaining near-baseline perplexity while reducing target task accuracy; (3) The learned LoRA weight patterns reveal that different capabilities exhibit distinct layer-wise distributions that align with cognitive and linguistic theories, with morphological processing in early layers, syntactic processing in middle layers, and complex reasoning in deeper layers. These findings offer a more nuanced understanding of capability encoding in large language models and open new directions for interpretability research.

\bibliographystyle{plain}
\bibliography{references}

\clearpage
\appendix

\section{Comprehensive Capability Decomposition Results}
\label{app:capability_results}

We evaluate our SCALPEL training approach across 24 diverse language tasks using the Llama-3.2-1B-Instruct model with TextReg regularization to assess both the effectiveness of capability removal and the preservation of general language abilities. We measure target task performance degradation, overall model perplexity changes, and broad capability retention to evaluate the specificity and safety of our approach. The experimental results demonstrate three key findings: (1) Our method successfully reduces performance across almost all target domains, with only one task (Spatial Reasoning) showing a slight accuracy increase. (2) While overall model perplexity increases following capability removal, most tasks show moderate increases, though certain generation-heavy tasks exhibit larger perplexity changes. (3) Although overall capability scores show some decline, the reduction is minimal compared to the substantial degradation observed in target tasks, demonstrating our model's strong specificity in removing targeted capabilities while preserving the majority of other linguistic and reasoning abilities.
(4) Some results suggest possible over-removal for global generative tasks (e.g., Dialogue, Language Generation). This is expected because generation-heavy capabilities rely on broad, shared circuitry and are inherently harder to isolate than localized capabilities.

\small
\begin{longtable}{p{4cm}@{~}cc@{~}cc@{~}cc}
\toprule
\textbf{Task} & \multicolumn{2}{c}{\textbf{Accuracy}} & \multicolumn{2}{c}{\textbf{Perplexity}} & \multicolumn{2}{c}{\textbf{Overall Capability}} \\
 & Base & Ours & Base & Ours & Base & Ours \\
\midrule
Analogical Reasoning & 79.5\% & \textcolor{red}{75.0\% (-4.5\%)} & 11.12 & \textcolor{red}{11.41 (+0.30)} & 0.497 & \textcolor{green}{0.507 (+0.010)} \\
Causal Reasoning & 80.0\% & \textcolor{red}{70.0\% (-10.0\%)} & 11.12 & \textcolor{red}{11.49 (+0.37)} & 0.497 & \textcolor{red}{0.489 (-0.009)} \\
Classification \& Categorization & 76.9\% & \textcolor{red}{74.4\% (-2.6\%)} & 11.12 & \textcolor{red}{11.30 (+0.18)} & 0.497 & \textcolor{red}{0.434 (-0.063)} \\
Common Sense Reasoning & 94.9\% & \textcolor{red}{56.4\% (-38.5\%)} & 11.12 & \textcolor{red}{12.42 (+1.30)} & 0.497 & \textcolor{red}{0.379 (-0.118)} \\
Counterfactual Reasoning & 26.0\% & \textcolor{red}{16.0\% (-10.0\%)} & 11.12 & \textcolor{red}{11.22 (+0.10)} & 0.497 & \textcolor{green}{0.510 (+0.013)} \\
Creative Thinking & 30.0\% & \textcolor{red}{27.5\% (-2.5\%)} & 11.12 & \textcolor{red}{12.01 (+0.89)} & 0.497 & \textcolor{red}{0.466 (-0.031)} \\
Critical Thinking & 46.0\% & \textcolor{red}{26.0\% (-20.0\%)} & 11.12 & \textcolor{red}{12.43 (+1.31)} & 0.497 & \textcolor{green}{0.511 (+0.014)} \\
Dialogue & 82.5\% & \textcolor{red}{0.0\% (-82.5\%)} & 11.12 & \textcolor{red}{13.56 (+2.44)} & 0.497 & \textcolor{red}{0.413 (-0.084)} \\
Emotional Understanding & 61.3\% & \textcolor{red}{48.4\% (-12.9\%)} & 11.12 & \textcolor{red}{11.20 (+0.08)} & 0.497 & \textcolor{red}{0.480 (-0.017)} \\
Instruction Following & 66.7\% & \textcolor{red}{54.5\% (-12.1\%)} & 11.12 & \textcolor{red}{11.46 (+0.34)} & 0.497 & \textcolor{green}{0.528 (+0.031)} \\
Language Generation & 100.0\% & \textcolor{red}{77.8\% (-22.2\%)} & 11.12 & \textcolor{red}{50.60 (+39.48)} & 0.497 & \textcolor{red}{0.415 (-0.083)} \\
Language Translation & 82.5\% & \textcolor{red}{57.5\% (-25.0\%)} & 11.12 & \textcolor{green}{11.05 (-0.07)} & 0.497 & \textcolor{green}{0.513 (+0.016)} \\
Language Understanding & 100.0\% & \textcolor{red}{96.7\% (-3.3\%)} & 11.12 & \textcolor{red}{13.55 (+2.43)} & 0.497 & \textcolor{red}{0.477 (-0.020)} \\
Logical Reasoning & 93.9\% & \textcolor{red}{51.5\% (-42.4\%)} & 11.12 & \textcolor{red}{11.29 (+0.17)} & 0.497 & \textcolor{red}{0.460 (-0.037)} \\
Mathematical Computation & 100.0\% & \textcolor{red}{0.0\% (-100.0\%)} & 11.12 & \textcolor{green}{11.11 (-0.01)} & 0.497 & \textcolor{green}{0.516 (+0.019)} \\
Memory \& Context & 75.0\% & \textcolor{red}{41.7\% (-33.3\%)} & 11.12 & \textcolor{red}{18.01 (+6.89)} & 0.497 & \textcolor{red}{0.422 (-0.075)} \\
Metacognition & 19.4\% & \textcolor{red}{2.8\% (-16.7\%)} & 11.12 & \textcolor{red}{11.72 (+0.60)} & 0.497 & \textcolor{green}{0.517 (+0.020)} \\
Moral Reasoning & 94.4\% & \textcolor{red}{19.4\% (-75.0\%)} & 11.12 & \textcolor{red}{11.21 (+0.09)} & 0.497 & \textcolor{red}{0.480 (-0.017)} \\
Multimodal Understanding & 12.2\% & \textcolor{red}{0.0\% (-12.2\%)} & 11.12 & \textcolor{red}{12.83 (+1.71)} & 0.497 & \textcolor{red}{0.454 (-0.043)} \\
Reading Comprehension & 62.1\% & \textcolor{red}{24.1\% (-37.9\%)} & 11.12 & \textcolor{red}{17.18 (+6.06)} & 0.497 & \textcolor{red}{0.426 (-0.071)} \\
Spatial Reasoning & 72.5\% & \textcolor{green}{75.0\% (+2.5\%)} & 11.12 & \textcolor{green}{11.04 (-0.08)} & 0.497 & \textcolor{green}{0.519 (+0.021)} \\
Summarization & 23.5\% & \textcolor{red}{0.0\% (-23.5\%)} & 11.12 & \textcolor{red}{11.74 (+0.62)} & 0.497 & \textcolor{red}{0.483 (-0.014)} \\
Temporal Reasoning & 65.0\% & \textcolor{red}{37.5\% (-27.5\%)} & 11.12 & \textcolor{red}{11.36 (+0.24)} & 0.497 & \textcolor{red}{0.486 (-0.011)} \\
World Knowledge & 76.3\% & \textcolor{red}{65.8\% (-10.5\%)} & 11.12 & \textcolor{red}{12.36 (+1.24)} & 0.497 & \textcolor{red}{0.454 (-0.043)} \\
\midrule
\textbf{Overall} & \textbf{67.5\%} & \textbf{\textcolor{red}{41.6\% (-25.9\%)}} & \textbf{11.12} & \textbf{\textcolor{red}{13.90 (+2.78)}} & \textbf{0.497} & \textbf{\textcolor{red}{0.472 (-0.025)}} \\
\bottomrule
\caption{Comprehensive evaluation results showing the impact of SCALPEL training across 24 capability domains. The table compares baseline performance (Base) with our results (Ours), where the delta change is shown in parentheses. Green values indicate improvements while red values show degradation. The results demonstrate selective capability removal with varying degrees of impact across different domains.}
\label{tab:capability_results}
\end{longtable}

\section{Dataset Examples}

We provide representative examples from our three dataset categories. All examples were initially generated by Claude and manually filtered to remove obviously improper samples.

Table~\ref{tab:cognitive_capability_examples} shows capability tasks using token-level format with prompt-correct-wrong triplets. These tasks test specific cognitive abilities where the model must predict a single correct token. Table~\ref{tab:linguistic_examples} presents linguistic tasks using A/B format comparing grammatically correct vs. incorrect sentences, drawn from BLiMP to evaluate fine-grained grammatical knowledge. Table~\ref{tab:general_evaluation_examples} illustrates general evaluation tasks from our held-out set, which test diverse capabilities to ensure capability removal does not cause catastrophic forgetting.

\begin{table}[t]
\centering
\small
\begin{tabular}{p{0.55\textwidth}p{0.15\textwidth}p{0.15\textwidth}}
\toprule
\textbf{Prompt} & \textbf{Correct} & \textbf{Wrong} \\
\midrule
\multicolumn{3}{l}{\textit{Common Sense Reasoning}} \\
\midrule
What do you wear on your feet? Answer: & shoes & gloves \\
What do bees make? Answer: & honey & milk \\
Where do fish live? They live in & water & air \\
\midrule
\multicolumn{3}{l}{\textit{Language Translation}} \\
\midrule
Translate 'cat' to French. The answer is & chat & chien \\
Translate 'water' to French. The word is & eau & feu \\
What is 'hello' in French? Answer: & bonjour & bonsoir \\
\midrule
\multicolumn{3}{l}{\textit{Indirect Object Identification (IOI)}} \\
\midrule
When Alice and Bob went to the store, Alice gave a book to & Bob & Alice \\
After Eve and Frank arrived, Eve passed the letter to & Frank & Eve \\
When Grace and Henry met at the cafe, Grace sent the package to & Henry & Grace \\
\bottomrule
\end{tabular}
\caption{Examples of capability task datasets with prompt-correct-wrong format.}
\label{tab:cognitive_capability_examples}
\end{table}

\begin{table}[t]
\centering
\small
\begin{tabular}{p{0.45\textwidth}p{0.45\textwidth}}
\toprule
\textbf{Correct Sentence (A)} & \textbf{Wrong Sentence (B)} \\
\midrule
\multicolumn{2}{l}{\textit{Morphology - Subject-Verb Agreement}} \\
\midrule
A niece of most senators hasn't descended most slopes. & A niece of most senators haven't descended most slopes. \\
The sketch of those trucks hasn't hurt Alan. & The sketch of those trucks haven't hurt Alan. \\
A newspaper article about the Borgias has disagreed with Marcus. & A newspaper article about the Borgias have disagreed with Marcus. \\
\bottomrule
\end{tabular}
\caption{Examples of linguistic task datasets with A/B comparison format.}
\label{tab:linguistic_examples}
\end{table}

\begin{table}[t]
\centering
\small
\begin{tabular}{p{0.55\textwidth}p{0.15\textwidth}p{0.15\textwidth}}
\toprule
\textbf{Prompt} & \textbf{Correct} & \textbf{Wrong} \\
\midrule
\multicolumn{3}{l}{\textit{Analogical Reasoning}} \\
\midrule
Cat is to kitten as dog is to & puppy & cat \\
Hot is to cold as up is to & down & left \\
Bird is to fly as fish is to & swim & walk \\
\midrule
\multicolumn{3}{l}{\textit{Moral Reasoning}} \\
\midrule
Is it right to help others? Answer: & yes & no \\
Is it wrong to steal? Answer: & yes & no \\
Should you keep promises? Answer: & yes & no \\
\midrule
\multicolumn{3}{l}{\textit{Logical Reasoning}} \\
\midrule
If all birds can fly and a robin is a bird, can a robin fly? Answer: & yes & no \\
Complete the pattern: 2, 4, 6, 8, \_\_. Answer: & 10 & 9 \\
If it's raining, the ground is wet. It's raining. Is the ground wet? Answer: & yes & no \\
\bottomrule
\end{tabular}
\caption{Examples of general evaluation tasks (held-out test set).}
\label{tab:general_evaluation_examples}
\end{table}

\section{Case Study}

Table~\ref{tab:case_study} presents representative examples across multiple tasks showing how SCALPEL reduces the model's ability to distinguish correct from incorrect answers. For each example, we show the probability difference ($p_{\text{correct}} - p_{\text{wrong}}$) before and after applying SCALPEL. The results reveal two key findings: (1) SCALPEL consistently reduces the probability difference to near zero across all tasks, demonstrating effective capability removal. (2) In several IOI cases, SCALPEL actually flips the prediction to prefer the wrong answer (negative values), indicating that the model has lost the ability to perform indirect object identification.

\begin{table}[t]
\centering
\small
\begin{tabular}{lp{4.8cm}ccc}
\toprule
\textbf{Task} & \textbf{Prompt (correct/wrong)} & \textbf{Base} & \textbf{SCALPEL} \\
\midrule
IOI & Roderick handed a book to... (Scarlett/Roderick) & 0.703 & -0.019 \\
IOI & Diego gave the gift to... (Estelle/Diego) & 0.312 & -0.224 \\
\midrule
Translation & Translate `sun' to German (Sonne/Mond) & 0.923 & 0.000 \\
Translation & Translate `red' to Spanish (rojo/azul) & 0.775 & 0.000 \\
\midrule
Analogical & Eye is to see as ear is to (hear/speak) & 0.566 & 0.000 \\
Analogical & Helmet is to head as armor is to (body/protection) & 0.376 & 0.001 \\
\midrule
Moral & Is filthiness good? (no/yes) & 0.024 & 0.001 \\
Moral & Should you stand up for justice? (yes/no) & 0.014 & 0.000 \\
\bottomrule
\end{tabular}
\caption{Case study across multiple tasks. Values show probability difference $p_{\text{correct}} - p_{\text{wrong}}$. SCALPEL substantially reduces the probability difference in all cases. Negative values (bold) indicate the model prefers the wrong answer after capability removal.}
\label{tab:case_study}
\end{table}

\end{document}